\documentclass[journal]{IEEEtran}

\usepackage[T1]{fontenc}
\usepackage[utf8]{inputenc}
\usepackage{graphicx}
\usepackage{amsmath}
\usepackage{subcaption}
\usepackage{cite}
\usepackage[hidelinks]{hyperref}

\title{Beyond Backscatter: AlphaEarth Land-Cover Priors for Rapid SAR Flood Segmentation Across Foundation Backbones}

\author{Sanjay Thasma, Yu-Hsuan Ho, and Ali Mostafavi%
\thanks{S. Thasma is with the Department of Computer Science and Engineering, Texas A\&M University, College Station, TX (e-mail: thasma@tamu.edu).}%
\thanks{Y.-H. Ho and A. Mostafavi are with the Urban Resilience.AI Lab, Zachry Department of Civil and Environmental Engineering, Texas A\&M University, College Station, TX (e-mail: yuhsuanho@tamu.edu; amostafavi@civil.tamu.edu).}%
\thanks{Corresponding author: Sanjay Thasma.}}

\begin{document}
\maketitle

\begin{abstract}
Rapid flood mapping is critical for emergency response, yet optical imagery is often unusable during major flooding and single-temporal SAR observations remain ambiguous because new inundation, permanent water, and other smooth surfaces can produce similar backscatter. This study evaluates whether stable land-context priors can improve post-event SAR flood segmentation when a registered, seasonally matched pre-event acquisition is unavailable. Using the CONUS (Continental United States) subset of ImpactMesh-Flood, we compare four backbones spanning distinct pretraining regimes---a from-scratch CNN UNet, an ImageNet-pretrained UNet, the SAR-pretrained TerraMind Vision Transformer, and the optical-satellite-pretrained DINOv3 Vision Transformer---in SAR-only, SAR+DEM, and SAR+AlphaEarth configurations under an identical fusion design, training protocol, and event-stratified split. Models are selected on a validation flood event and evaluated separately on two held-out events, Hurricane Florence and the Louisiana floods, with three-seed reporting for auxiliary configurations. Both auxiliary priors improve over the observed SAR-only baselines across all backbones and test events. AlphaEarth exceeds DEM on the harder Florence event for every backbone and achieves the best Florence IoU, while DEM is competitive on Louisiana and produces the best result there. The seed analysis reveals an important trade-off: DEM is generally more stable across initializations, whereas AlphaEarth can provide higher peak performance and a higher-recall operating point on the more difficult event. Cross-event differences are strongly associated with flood-class prevalence and similarity to the training distribution, underscoring the need for per-event evaluation. The study reframes single-temporal SAR flood segmentation as an alignment problem between event-specific radar observations and stable land-surface priors, showing that learned and physical auxiliary context provide complementary pathways for more reliable rapid flood mapping.
\end{abstract}


\IEEEpeerreviewmaketitle

\section{Introduction}
Floods rank among the most frequent and economically destructive natural hazards worldwide, and the speed with which inundation extent is mapped after an event directly shapes evacuation routing, search-and-rescue prioritization, and the activation of recovery funding, and it feeds downstream operational intelligence such as near-real-time flood-damage nowcasting~\cite{liu2024flooddamagecast}. Optical Earth observation has long served as the default mapping modality~\cite{xiao2025damagecat, kaur2023large}. However, the dense cloud cover that typically accompanies major flooding obscures the surface precisely when intelligence is most needed~\cite{ho2025multimodal, wang2024scalable}. Synthetic Aperture Radar (SAR) circumvents this barrier through its all-weather, day-night sensing capability, and Sentinel-1 in particular has become the operational backbone of large-scale flood-extent products~\cite{singh2021review,amitrano2024flood}.

The operational value of flood segmentation depends on performance under the exact conditions in which mapping is most difficult: incomplete situational information, no guarantee of a matched pre-event scene, cloud-obscured optical imagery, and rapidly changing flood dynamics. SAR provides the essential all-weather observation channel, but a single post-event SAR image is not self-explanatory. Low backscatter can indicate newly inundated land, permanent water, smooth wet surfaces, radar shadow, or built-environment effects. This ambiguity is why many high-performing flood-mapping workflows rely on some form of auxiliary information, whether a pre-event image, a permanent-water mask, topographic context, or multi-sensor time series. The unresolved question is not simply which backbone produces the highest IoU, but which external priors make SAR observations more interpretable and transferable across events.

This gap is increasingly important as geospatial foundation models and large embedding products become available for disaster applications. These models promise reusable land-surface knowledge, but their value for event-specific hazard mapping should be tested against simpler and cheaper baselines rather than assumed. DEM provides a physically meaningful and globally available prior on local topography; AlphaEarth provides a learned, high-dimensional representation of land-cover and surface context. The central question for rapid flood mapping is therefore whether a learned land-cover prior adds operationally meaningful context beyond topography, and whether that benefit persists across architectures, events, and random initializations. Answering this question requires an evaluation design that separates the contribution of the auxiliary signal from the contribution of the segmentation backbone.

Established SAR flood-mapping pipelines fall into two families, each with characteristic failure modes. Bi-temporal change detection compares a pre-event scene against the post-event acquisition to suppress permanent water, vegetation, and built-up surfaces~\cite{saleh2024high, ho2025multimodal, misra2025mapping}. Recent bi-temporal architectures, including attention-based Siamese networks~\cite{yadav2022adsu} and differential-attention vision transformers~\cite{saleh2023damnet}, continue to demonstrate the value of temporal differencing for suppressing permanent water on multi-temporal Sentinel-1 data. The approach is highly sensitive to seasonal and short-term environmental shifts between the two acquisition dates, vegetation phenology, soil moisture, snow cover, and crop state can change appreciably within the days or weeks separating the two scenes~\cite{zwieback2015assessment, eshqi2018band, zhang2018analysis}. Single-temporal post-event detection avoids these constraints by classifying water directly from the post-event scene alone~\cite{bereczky2022sentinel}, and label-scarce variants of this setting have motivated semi-supervised strategies such as confidence-filtered pseudo-labeling~\cite{paul2021semisupervised}. However, it loses the radiometric anchor that permanent water provides under change detection, and consequently struggles to distinguish floodwaters from rivers, lakes, reservoirs, and coastal water whose backscatter signatures overlap with inundation~\cite{nagai2021sar}. Both families therefore depend on an auxiliary signal that is brittle in practice, whether a registered, seasonally matched pre-event acquisition or an external means of identifying permanent water.

Methodological development in flood segmentation has been shaped as much by the benchmark datasets used for training and evaluation as by advances in model architecture. The evolution of these datasets generally follows a progression from single-temporal paradigms to multi-temporal radar and, ultimately, dense multimodal time series. Early benchmarks established baselines by focusing on single, post-event acquisitions to detect floodwater. For instance, Sen1Floods11 provides Sentinel-1 SAR in VV and VH polarizations and Sentinel-2 optical imagery, utilizing both automated remote sensing thresholding and manual annotations for validation~\cite{bonafilia2020sen1floods11}. Similarly focused on post-event mapping, ETCI-2021 provides Sentinel-1 SAR imagery in VV and VH polarizations alongside binary flood masks. To suppress the permanent-water ambiguity inherent in this single-scene analysis, subsequent datasets framed flood detection fundamentally as a change-detection problem using time-series radar data~\cite{ETCI2021}. UrbanSARFloods pushes this paradigm into densely built-up environments by utilizing pre-event, co-event, and post-event Sentinel-1 Single Look Complex (SLC) data to leverage both intensity and interferometric coherence, employing a hybrid labeling strategy that combines semi-automatic techniques with manual annotations for test subsets~\cite{UrbanSARFloods}. Expanding this multi-temporal radar approach globally, Kuro Siwo provides triplets of Sentinel-1 imagery in both GRD and SLC formats, paired with Digital Elevation Model (DEM) data, featuring labels created entirely through manual photo interpretation~\cite{bountos2023kuro}. Recognizing the limitations of relying solely on radar, the most recent datasets bridge the optical-SAR divide. Unlike earlier datasets that used optical imagery primarily to generate training labels, these newer benchmarks supply densely co-registered multi-sensor time series. SEN12-FLOOD, an early multimodal benchmark, supplies sequences of Sentinel-1 SAR and Sentinel-2 optical imagery with annotations sourced from the Copernicus Emergency Management Service (EMS)~\cite{SEN12-FLOOD}. More recently, ImpactMesh-Flood seamlessly integrates Sentinel-1 SAR, Sentinel-2 optical imagery, and Copernicus DEM data across four temporal observations per event, opening the door to auxiliary-context approaches~\cite{ImpactMeshFlood}. Ultimately, across this progression, the choice of benchmark has reinforced rather than bridged the bi-temporal versus single-temporal divide. Datasets with paired acquisitions naturally favor change-detection pipelines, while single-scene datasets surface the permanent-water ambiguity that change detection suppresses. Auxiliary signals capable of compensating for the failure modes of both families without requiring a registered pre-event acquisition have received comparatively little systematic study. While the inclusion of DEM in recent datasets represents an initial attempt at providing auxiliary signals, elevation data alone fails to capture the broader environmental and land-surface representations necessary to fully resolve these ambiguities. This underscores the critical need for a temporally stable, dataset-agnostic land cover prior to disambiguate SAR observations.

A parallel development to the evolution of benchmark datasets has been the emergence of large vision foundation models, which introduce markedly different inductive biases to the flood segmentation task. Vision Transformers (ViT)~\cite{dosovitskiy2020image} pretrained through self-supervised objectives produce robust features that transfer effectively to a wide range of downstream segmentation tasks. Notably, recent foundation models like DINOv3~\cite{simeoni2025dinov3} have expanded their pretraining distributions beyond massive natural-image corpora to explicitly include optical satellite imagery. However, while this incorporates valuable Earth observation context, its strictly optical pretraining distribution still remains far removed from the speckle-dominated radiometry characteristic of SAR imagery. Conversely, multimodal geospatial foundation models, such as TerraMind~\cite{jakubik2025terramind}, attempt to directly close this domain gap by pretraining on a diverse, aligned set of Earth observation modalities, explicitly integrating both SAR and optical scenes alongside auxiliary geospatial data. This builds on a broader line of Earth-observation self-supervised pretraining, including masked-autoencoder approaches for temporal and multi-spectral imagery~\cite{cong2022satmae}, scale-aware variants~\cite{reed2022scalemae}, and generalist geospatial foundation models such as Prithvi~\cite{jakubik2023prithvi}, which have begun to be evaluated directly on flood inundation mapping~\cite{li2023geoai} and extended to sensor-flexible architectures that accept SAR, multispectral, or fused inputs within a single model~\cite{tanaka2025sensor}. The relative performance hierarchy of these backbones in operational flood mapping is not yet settled, and it remains unclear how much of the performance gap between them reflects raw architectural capacity versus pretraining-domain alignment.

Beyond the choice of backbone, land cover has long been recognized as a valuable auxiliary signal for flood mapping, typically incorporated through SAR-optical fusion or Digital Elevation Models. AlphaEarth (AE) embeddings represent a more general and scalable alternative: a 64-dimensional, per-pixel land cover code derived from a variational model trained on multimodal Earth observation data, available as a raster aligned to standard satellite grids~\cite{brown2025alphaearth}. Because AE embeddings are temporally aggregated rather than tied to a specific acquisition date, they are largely insensitive to the seasonal and short-term environmental shifts that destabilize bi-temporal change detection. This view aligns with work treating flood-relevant context as an event-independent property of place rather than a single observation, for example interpretable models of property-level flood predisposition~\cite{liu2024floodgenome}. Simultaneously, they may encode land-cover context relevant to permanent water, which may supply disambiguation context that single-temporal detectors otherwise lack. AE therefore serves as a candidate prior for the failure modes of both pipeline families. Yet most evaluations do not isolate whether such a learned land-surface representation provides additional value over a simple physical prior, nor whether that value is consistent across backbones with different pretraining histories, so its empirical worth as auxiliary context remains unmeasured. As its counterpoint we adopt DEM, a minimal, physically meaningful, and globally available single-channel prior on local topography, against which the value of AE's richer, higher-dimensional encoding can be tested. Departing from architecture-centric flood-mapping studies, we treat post-event SAR segmentation as an alignment problem between an event-specific radar observation and a stable representation of the underlying land surface: holding the fusion design and training protocol fixed, we vary only two interpretable axes, the backbone pretraining regime and the auxiliary signal, across from-scratch, ImageNet-pretrained, SAR-pretrained, and optical-satellite-pretrained backbones under an event-stratified split. This makes the study a controlled test of prior utility rather than another model leaderboard, addressing a specific empirical gap: when does a learned land-cover prior improve cross-event generalization enough to justify its additional dimensionality and seed sensitivity?

Concretely, the study pursues three objectives: first, to determine whether stable land-context priors improve single-temporal post-event SAR flood segmentation across backbones with different pretraining histories; second, to test whether a learned land-surface prior (AlphaEarth) adds value beyond a simpler physical prior (DEM); and third, to characterize how these priors affect cross-event generalization, seed stability, and precision-recall behavior under an event-stratified protocol.

The contributions of this work are threefold: (i) a controlled comparison of four SAR backbones representing distinct pretraining regimes, a from-scratch CNN UNet~\cite{ronneberger2015unet}, an ImageNet-pretrained UNet with a ResNet-50 backbone~\cite{bonafilia2020sen1floods11, he2016resnet}, a SAR-pretrained Vision Transformer (TerraMind)~\cite{jakubik2025terramind}, and an optical-satellite-pretrained Vision Transformer (DINOv3)~\cite{simeoni2025dinov3}, each evaluated in SAR-only, SAR--DEM, and SAR--AlphaEarth configurations under an identical fusion design, optimizer, loss function, and event-stratified data split with per-event test reporting; (ii) empirical evidence that both auxiliary signals improve on the SAR-only baseline across all four architectures, with AlphaEarth exceeding the simpler DEM prior on the harder cross-event test for every backbone, while the two are competitive on the easier event; and (iii) a seed-stability analysis showing that DEM is the more stable signal across random initializations whereas AlphaEarth attains higher peak performance on the harder event, together with a measured account of the cross-event performance gap in terms of flood-class prevalence. The remainder of this paper details the dataset, proposed fusion design, and training protocol; presents the experimental results on the ImpactMesh-Flood continental United States (CONUS) subset; offers an interpretive discussion alongside study limitations; and closes with concluding remarks.

\section{Methodology}

\subsection{Data}
The study uses the ImpactMesh-Flood dataset published by IBM, DLR, and the ESA $\Phi$-lab on HuggingFace~\cite{ImpactMeshFlood}, focusing exclusively on the CONUS subset spanning seven flood events: EMSR229 (Hurricane Harvey, Texas), EMSR203 (floods in central United States), EMSR186 (Hurricane Matthew, southeastern United States), EMSR311 (Hurricane Florence, US East Coast), EMSR176 (Louisiana floods), EMSR322 (Hurricane Michael, Florida/Alabama/Georgia), and EMSR241 (Hurricane Irma, Florida). These events span a mix of urban and rural flood scenarios, presenting a challenging generalization setting for post-event SAR flood segmentation. The dataset provides post-event Sentinel-1 Radiometric Terrain Correction (RTC) imagery in VV and VH polarizations at 10m spatial resolution alongside flood extent masks derived from Copernicus EMS activation polygons. The full CONUS subset contains 10,197 tiles of $256 \times 256$ pixels with each tile covering $2.56 \times 2.56$ km on the ground.

To eliminate spatial data leakage and evaluate true cross-event generalization, all experiments use an event-stratified split. Training tiles are drawn from three Gulf and Atlantic Coast hurricane events (EMSR229 Harvey, EMSR186 Matthew, EMSR322 Michael) totaling 6,676 tiles. Validation tiles are drawn from EMSR203 (Central US floods, 2,384 tiles) and used for model selection. Two events are held out as separate test sets: EMSR311 (Hurricane Florence, 756 tiles) and EMSR176 (Louisiana floods, 336 tiles), evaluated and reported independently to surface cross-event performance variability that aggregate metrics would obscure. EMSR241 (Hurricane Irma, 45 tiles) is excluded as too small for stable test metrics.

AlphaEarth(AE) embeddings are 64-dimensional, per-pixel land cover representations derived from a variational model trained on multimodal satellite data~\cite{brown2025alphaearth}. Each tile has a corresponding AE embedding of shape $(64, 256, 256)$. To make the temporal provenance of the AE prior explicit and to rule out the possibility that the embedding encodes information from the target flood event, we record the following acquisition details. The embeddings are drawn from a single annual AlphaEarth Foundations Satellite Embedding layer (Earth Engine collection \texttt{GOOGLE/SATELLITE\_EMBEDDING/V1/ANNUAL}, product version V1) for the 2024 reference year, which aggregates multi-sensor observations over the 2024 calendar year; the same layer is applied uniformly to all tiles and events, with no per-event year selection. The source embeddings, in EPSG:4326 and float64, are warped onto each tile's native CRS, affine transform, and $256\times256$ grid by bilinear resampling for near 1:1 co-registration with the 10\,m Sentinel-1 imagery, and stored as float16. The source declares no no-data value, and because AlphaEarth provides full CONUS land and shallow-water coverage, we applied no additional no-data masking. Because the 2024 reference year postdates every flood event (2016--2018) by six to eight years, the embeddings cannot encode the target inundation, eliminating event leakage, at the cost of reflecting contemporary rather than at-event land conditions. These embeddings encode a temporally aggregated baseline of the land surface, reflecting temporally aggregated land-surface context rather than a single pre-event acquisition, which remains robust to seasonal phenology and short-term environmental variation. Serving as auxiliary context to compensate for the absence of pre-event SAR imagery, they are insensitive to the seasonal and registration constraints of bi-temporal change detection, and potentially provide land-cover context relevant to the permanent-water ambiguity of single-temporal post-event detection. AE embeddings are normalized using per-pixel L2 normalization to preserve their unit sphere geometry prior to being passed to the model.

To benchmark AE against a simpler, widely available topographic signal, we additionally evaluate fusion with Copernicus Digital Elevation Model (DEM) data bundled with ImpactMesh-Flood. Each tile has a corresponding DEM raster of shape (256, 256) encoding ground elevation in meters on the same 10m grid as the SAR imagery. DEM tiles are normalized per-tile by subtracting the per-tile minimum and dividing by the per-tile range, yielding a [0, 1] representation of relative within-tile topography. Flat tiles with zero per-tile range (a constant elevation, giving a zero denominator) are mapped to an all-zero tile rather than divided, avoiding a division-by-zero. This per-tile normalization is selected because flood pooling depends on local elevation minima rather than absolute elevation. Prioritizing local relief over absolute elevation is appropriate for tile-level inundation, though absolute-elevation normalization is a reasonable alternative that could be tested in future work. DEM serves as the natural minimal baseline against which to evaluate whether the 64-dimensional AE embedding provides information beyond simple topography. Elevation and topographic descriptors are well established as physically meaningful flood signals, whether through lowest-floor elevation estimation for depth and damage modeling~\cite{ho2023elevvision} or anticipatory flood-warning frameworks built on height-above-nearest-drainage and distance-to-stream features~\cite{li2025flashflood}, which motivates DEM as a strong yet inexpensive prior.

SAR inputs are normalized following the Sen1Floods11 convention~\cite{bonafilia2020sen1floods11}. VV backscatter is clipped to the range $[-23, 0]$ dB and normalized with channel-wise mean 0.6851 and standard deviation 0.0820. VH backscatter is clipped to $[-28, -5]$ dB and normalized with mean 0.5235 and standard deviation 0.1102. For the CNN UNet and ImageNet-UNet, a third input channel is computed as the VV$-$VH ratio band, clipped to $[-20, 20]$ dB and instance-normalized per tile, as this ratio is a strong discriminator of open water. TerraMind and DINOv3 receive only the two-channel VV$+$VH input, as TerraMind was pretrained on two-channel Sentinel-1 RTC data and DINOv3 uses a trainable SAR adapter to map two-channel SAR input to the three-channel format expected by its ViT patch embedding.

\subsection{Approach}
The study tests the hypothesis that AlphaEarth land cover embeddings provide useful auxiliary context for single-temporal post-event SAR flood segmentation when pre-event imagery is unavailable. By supplying a temporally stable, co-registered land cover prior, these embeddings are intended to provide land-cover context relevant to distinguishing permanent water from active inundation, while remaining insensitive to the seasonal and short-term environmental shifts.

\begin{figure*}[!t]
\centering
\includegraphics[width=\textwidth]{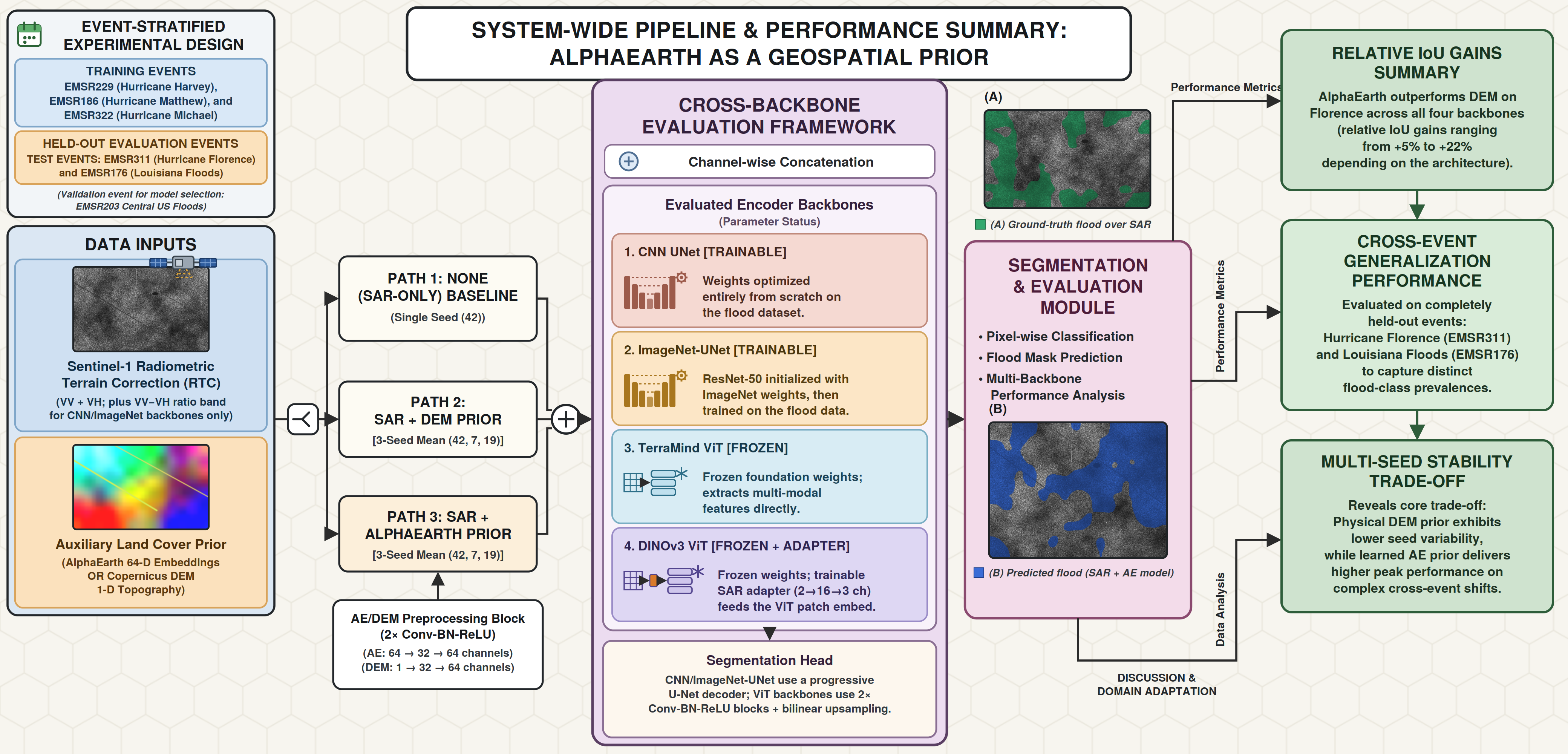}
\caption{System-wide pipeline and performance summary. Four SAR backbones spanning distinct pretraining regimes (from-scratch CNN UNet, ImageNet-pretrained UNet, frozen TerraMind ViT, and frozen DINOv3 ViT with a trainable SAR adapter) are each evaluated in SAR-only, SAR+DEM, and SAR+AlphaEarth configurations under an identical fusion design and event-stratified split. The AlphaEarth visualization shows the first three PCA components as RGB; example ground-truth and predicted flood masks are shown for a tile from EMSR229 (Hurricane Harvey).}
\label{fig:pipeline}
\end{figure*}

To evaluate this, we developed a multimodal segmentation pipeline. Figure \ref{fig:pipeline} illustrates the proposed framework, outlining the parallel data ingestion, normalization, dual-branch feature extraction, and multimodal fusion. Within this architecture, all models share a common fusion design, the SAR backbone produces a spatial feature map passed to a classification head. In SAR+AE variants, AE embeddings are processed through a lightweight AE branch consisting of two Conv-BN-ReLU blocks ($64 \rightarrow 32 \rightarrow 64$ channels) and bilinearly interpolated to match the backbone feature resolution. The resulting AE features are concatenated with the backbone features along the channel dimension. In SAR+DEM variants, the DEM raster is processed through an architecturally identical branch with the only difference being the input channel count ($1 \rightarrow 32 \rightarrow 64$), enabling a controlled comparison between the two auxiliary signals. The classification head is a $1\times1$ convolution producing two-class logits (flood/non-flood). For the ViT backbones (TerraMind, DINOv3), this is preceded by two Conv-BN-ReLU blocks at 256 channels and bilinear upsampling to input resolution. For the CNN UNet and ImageNet-UNet, the encoder-decoder already provides dense pixel features, so the $1\times1$ conv is applied directly. Four backbone architectures are evaluated, each tested in SAR-only, SAR+DEM, and SAR+AE configurations, for twelve configurations in total.

\textbf{CNN UNet}~\cite{ronneberger2015unet} is a standard encoder-decoder UNet trained from scratch with four encoding blocks ($3 \rightarrow 64 \rightarrow 128 \rightarrow 256 \rightarrow 512$ channels), a bottleneck with Dropout2d(0.3), and four symmetric decoding blocks with skip connections. This serves as the primary baseline because it carries no pretraining bias of any kind --- every parameter learns SAR flood segmentation directly from the ImpactMesh data.

\textbf{ImageNet-pretrained UNet} replicates the architectural and input configuration introduced in the Sen1Floods11 work~\cite{bonafilia2020sen1floods11}, which established a fully convolutional baseline for Sentinel-1 SAR flood segmentation across 11 global flood events. We adopt the same ResNet-50 encoder~\cite{he2016resnet} initialized with ImageNet weights, loaded via segmentation-models-pytorch~\cite{Iakubovskii:2019} with a UNet-style decoder, and use the same VV/VH channel-wise normalization statistics. The model is trained from scratch on ImpactMesh CONUS data rather than using a Sen1Floods11-pretrained checkpoint, so its only prior exposure is the ImageNet initialization of the encoder. While this represents a standard transfer learning baseline, the ImageNet pretraining domain, RGB natural images, introduces a known mismatch with SAR backscatter data that we examine in the discussion.

\textbf{TerraMind}~\cite{jakubik2025terramind} uses the \texttt{terramind\_v1\_base} Vision Transformer backbone, the first any-to-any generative multimodal foundation model for Earth observation, jointly developed by IBM, ESA, and Forschungszentrum J\"{u}lich. TerraMind is pretrained on 500B tokens from 9M spatiotemporally aligned multimodal samples across nine geospatial modalities including Sentinel-1 RTC, making it the only backbone in our evaluation with direct prior exposure to SAR backscatter data. The backbone is frozen throughout training and loaded via terratorch~\cite{gomes2025terratorch}. Only the segmentation head and AE branch are trained. Token outputs from the final transformer layer are reshaped to a spatial grid and bilinearly upsampled to $256 \times 256$.

\textbf{DINOv3}~\cite{simeoni2025dinov3} uses the \texttt{facebook/\allowbreak dinov3-\allowbreak vitl16-\allowbreak pretrain-\allowbreak sat493m} Vision Transformer loaded via HuggingFace Transformers. For this architecture, we specifically employ the variant pretrained exclusively on SAT-493M, a satellite dataset of 493 million RGB ortho-rectified Maxar images at 0.6m resolution that contains only optical data and no SAR exposure. The backbone is frozen throughout training. A trainable SAR adapter ($2 \rightarrow 16 \rightarrow 3$ channel convolution) maps the two-channel SAR input to the three-channel format expected by the ViT~\cite{dosovitskiy2020image} patch embedding. Patch tokens from the final layer are reshaped and upsampled identically to TerraMind.

All models are trained with AdamW ($\text{lr}=1\text{e-}4$, $\text{weight\_decay}=1\text{e-}2$) and a ReduceLROnPlateau scheduler (factor$=0.5$, patience$=4$, mode$=\max$). The loss function is a weighted sum of cross-entropy and Dice loss:
\begin{equation}
    \mathcal{L} = \mathcal{L}_{\text{CE}} + 0.5 \times \mathcal{L}_{\text{Dice}}
\end{equation}
Cross-entropy uses class weights $[1.0,\ 19.2]$, derived from the empirical background-to-flood pixel ratio measured across 2{,}000 randomly sampled training tiles, to address the severe class imbalance. Training runs for a maximum of 60 epochs with early stopping triggered after 10 consecutive epochs without improvement in validation IoU. Model selection (checkpoint saving), learning rate scheduling, and early stopping are all based on validation IoU rather than validation loss, so that the model state evaluated on the test sets is the one that performed best on the validation event by the same metric reported as the headline result. Batch size is 8 for SAR-only and 4 for SAR+AE and SAR+DEM loaders. Data augmentation consists of random horizontal and vertical flips applied identically to SAR inputs, auxiliary inputs (AE or DEM), and ground-truth masks. All random number generators are initialized to seed 42 for the primary runs. To characterize seed-dependent variance, each of the eight auxiliary-signal configurations (four SAR+DEM and four SAR+AE) is additionally trained under two further seeds (7 and 19), yielding three seeds per configuration. The four SAR-only baselines are trained under the primary seed only; multi-seed characterization of the baselines is left to future work.

\section{Results}

All experiments were run on an NVIDIA A100 GPU. The dataset split, normalization, loss function, optimizer, scheduler, augmentation, AE branch architecture, and DEM branch architecture were held constant across all twelve experiments. The only variables were the SAR backbone and the auxiliary signal configuration (SAR-only, SAR+DEM, or SAR+AE). This controlled design isolates the contribution of each auxiliary signal from architectural differences.

\subsection{Evaluation Protocol}
\label{sec:eval}

Models are selected on the validation event (EMSR203, Central US floods) by validation flood-class IoU. The checkpoint achieving the highest validation IoU is evaluated once on each of the two held-out test events, EMSR311 (Hurricane Florence, $n=756$) and EMSR176 (Louisiana floods, $n=336$), with no further checkpoint selection on the test sets. Results are reported per event rather than aggregated, to surface cross-event performance variability that a single aggregate metric would obscure.

All metrics are computed from globally aggregated true-positive, false-positive, and false-negative counts across all valid pixels in each evaluation set, where validity excludes pixels labeled with the dataset's ignore index (mask $= -1$). Validation and test metrics use the same aggregation, so the value used for model selection on the validation event is directly comparable to the value reported as the test-set headline. Flood-class IoU and F1 are defined as:
\begin{equation}
    \text{IoU} = \frac{TP}{TP + FP + FN}
\end{equation}
\begin{equation}
    \text{F1} = \frac{2 \cdot TP}{2 \cdot TP + FP + FN}
\end{equation}
where $TP$, $FP$, and $FN$ denote true positives, false positives, and false negatives respectively, computed over the flood class only. Per-event precision and recall, defined as $P = TP/(TP+FP)$ and $R = TP/(TP+FN)$, are additionally reported to characterize the operational trade-offs of each configuration. Unless stated otherwise, every reported metric for an auxiliary-signal configuration is the mean over three random seeds, 42, 7, and 19, and every reported SAR-only metric is the single primary-seed value from seed 42. Where individual seed-42 values are quoted for illustration, such as the precision and recall operating points, they are marked as seed 42 inline. Complete per-seed results for all configurations, including F1, precision, and recall on both events, are provided in Appendix~\ref{sec:appendix}.

\begin{table}[!t]
\centering
\small
\caption{Per-event test-set performance (flood-class IoU) for all twelve configurations on the held-out events EMSR311 (Hurricane Florence, $n=756$) and EMSR176 (Louisiana floods, $n=336$). Auxiliary configurations report the mean and standard deviation over three seeds, 42, 7, and 19, and SAR-only baselines are single-seed values from seed 42. Bold indicates the best mean IoU per event.}
\label{tab:results}
\setlength{\tabcolsep}{5pt}
\begin{tabular}{lcc}
\hline
\textbf{Configuration} & \textbf{Florence IoU} & \textbf{Louisiana IoU} \\
\hline
CNN UNet SAR-only       & 0.041            & 0.103 \\
CNN UNet SAR+DEM        & $0.046 \pm 0.006$ & $\mathbf{0.198 \pm 0.008}$ \\
CNN UNet SAR+AE         & $0.056 \pm 0.016$ & $0.167 \pm 0.007$ \\
\hline
ImageNet-UNet SAR-only  & 0.057            & 0.136 \\
ImageNet-UNet SAR+DEM   & $0.058 \pm 0.003$ & $0.156 \pm 0.013$ \\
ImageNet-UNet SAR+AE    & $0.061 \pm 0.015$ & $0.165 \pm 0.007$ \\
\hline
TerraMind SAR-only      & 0.046            & 0.144 \\
TerraMind SAR+DEM       & $0.064 \pm 0.009$ & $0.162 \pm 0.012$ \\
TerraMind SAR+AE        & $0.072 \pm 0.005$ & $0.161 \pm 0.010$ \\
\hline
DINOv3 SAR-only         & 0.045            & 0.133 \\
DINOv3 SAR+DEM          & $0.065 \pm 0.003$ & $0.164 \pm 0.016$ \\
DINOv3 SAR+AE           & $\mathbf{0.078 \pm 0.003}$ & $0.172 \pm 0.006$ \\
\hline
\end{tabular}
\end{table}

\subsection{Overall Performance Across All Configurations}

Table~\ref{tab:results} reports per-event test performance, averaged over three seeds, for all twelve configurations. Several patterns are visible.

First, auxiliary signals consistently help over SAR-only baselines. Averaged over seeds, both DEM and AE improve on SAR-only baseline across all four backbones on both test events, indicating that a single post-event SAR scene alone is insufficient for cross-event flood segmentation.

Second, no single configuration dominates both test events. DINOv3 SAR+AE achieves the best Florence IoU ($0.078$), while CNN UNet SAR+DEM achieves the best Louisiana IoU ($0.198$). The best configuration thus depends on the event.

Third, AlphaEarth is directionally stronger than DEM on Florence, the harder of the two test events, across all four backbones, though several per-backbone margins fall within seed uncertainty and the small number of seeds ($n=3$) makes these differences indicative rather than confirmed. The mean Florence relative IoU gain of AE over DEM ranges from $+5\%$ (ImageNet-UNet) to $+22\%$ (CNN UNet). On the easier Louisiana event the two signals are competitive: AE leads on ImageNet-UNet and DINOv3, while DEM leads on CNN UNet and is essentially tied on TerraMind. The relationship between this pattern, backbone pretraining, and seed sensitivity is examined in Sections~\ref{sec:per-event} and~\ref{sec:seed-stability}.

Fourth, absolute IoU remains low across all configurations, ranging from $0.04$ to $0.20$. This reflects three factors: the difficulty of cross-event single-temporal SAR flood segmentation, the coarseness of Copernicus EMS labels (discussed in Section~\ref{sec:limitations}), and the domain shift between the hurricane-driven training events and the more inland flooding in the test events.

\begin{figure*}[!t]
\centering
\includegraphics[width=\textwidth]{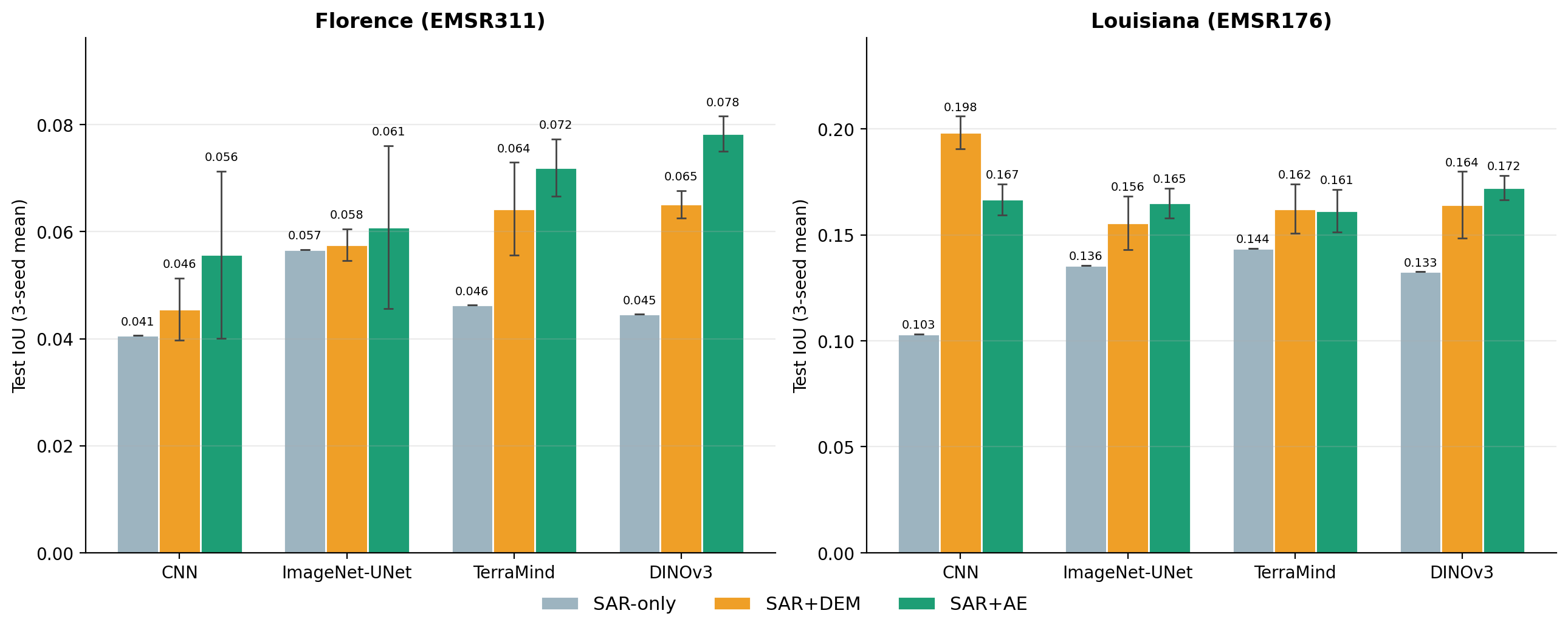}
\caption{Test IoU by backbone and auxiliary configuration for the two held-out events, averaged over three seeds (error bars show standard deviation; SAR-only baselines are single-seed). AlphaEarth exceeds DEM on Florence across all four backbones; the two signals are competitive on Louisiana.}
\label{fig:test_iou_by_config}
\end{figure*}

\subsection{Per-Event Test Performance}
\label{sec:per-event}

Florence and Louisiana differ substantially in absolute IoU magnitude. Across the twelve configurations, mean Florence IoU ranges from $0.04$ to $0.08$, while mean Louisiana IoU ranges from $0.10$ to $0.20$. Every configuration achieves higher IoU on Louisiana than on Florence. Possible causes are discussed in Section~\ref{sec:cross-event}.

On Florence, the best mean IoU is $0.078$ (DINOv3 SAR+AE), followed by TerraMind SAR+AE ($0.072$), DINOv3 SAR+DEM ($0.065$), and TerraMind SAR+DEM ($0.064$). AlphaEarth is the stronger auxiliary signal on this event: averaged over seeds, AE exceeds DEM for all four backbones, and the two best-performing configurations are both SAR+AE on frozen-backbone foundation models.

On Louisiana, the best mean IoU is $0.198$ (CNN UNet SAR+DEM), followed by DINOv3 SAR+AE ($0.172$), ImageNet-UNet SAR+AE ($0.165$), and DINOv3 SAR+DEM ($0.164$). The two signals are competitive on this event: AE leads on ImageNet-UNet and DINOv3, DEM leads on CNN UNet, and the two are within one standard deviation on TerraMind. The single strongest Louisiana result, CNN UNet SAR+DEM, pairs the simplest auxiliary signal with the only fully-from-scratch backbone.

The two auxiliary signals produce different precision-recall operating points. SAR+AE configurations tend toward higher recall: ImageNet-UNet SAR+AE achieves recall $0.824$ on Louisiana at precision $0.164$ (seed 42). SAR+DEM configurations tend toward higher precision: CNN UNet SAR+DEM on Louisiana reaches precision $0.221$ at recall $0.658$ (seed 42). The single-seed values (seed 42) are illustrative of this trend rather than averaged summary statistics; full three-seed precision and recall means are reported in Appendix~\ref{sec:appendix}. AE-paired models are more inclusive in their flood predictions; DEM-paired models are more conservative. The two operating points suit different applications, depending on whether missed detections or false alarms are more costly.

\subsection{Seed Stability}
\label{sec:seed-stability}

\begin{figure*}[!t]
\centering
\includegraphics[width=\textwidth]{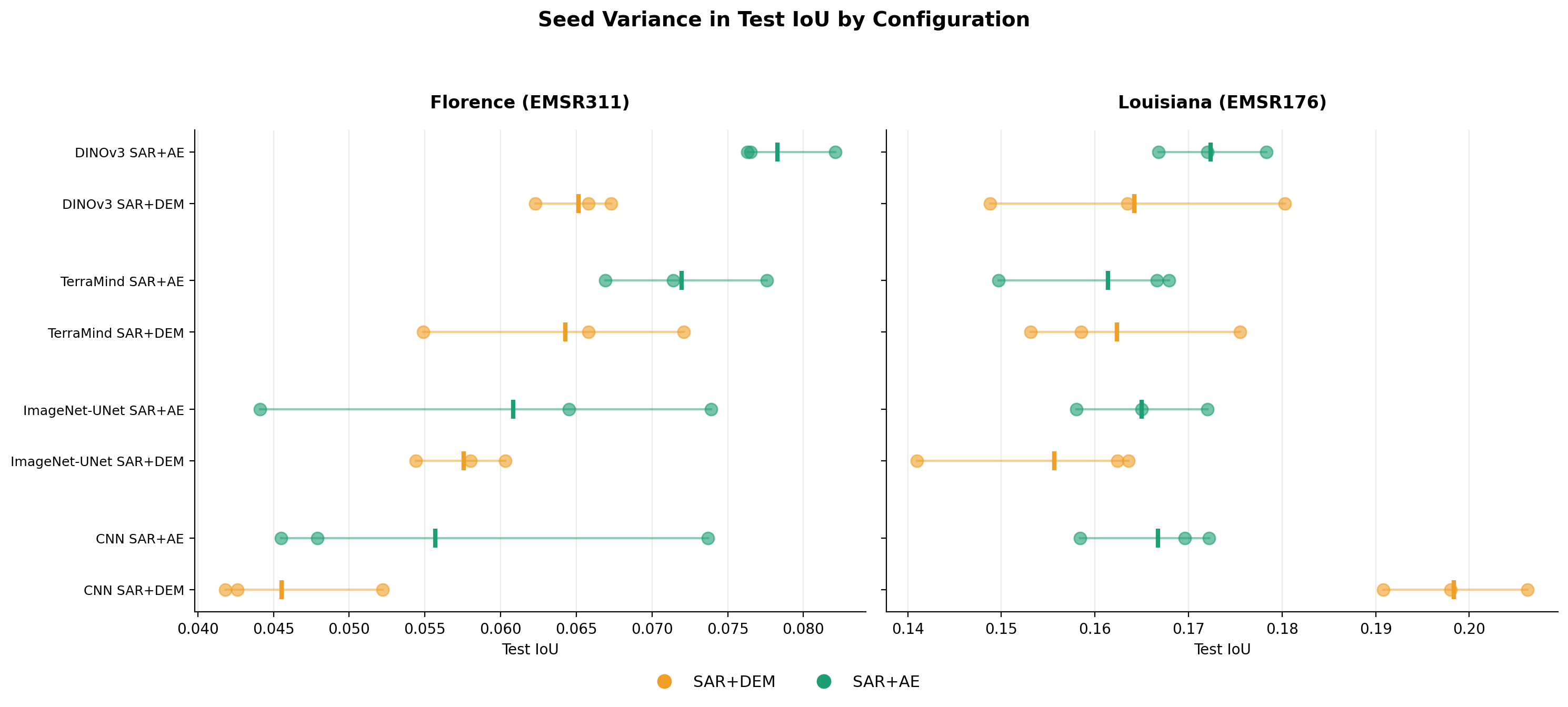}
\caption{Test IoU across three seeds for each auxiliary configuration, on both held-out events. Dots mark individual seeds; the vertical bar marks the mean. On Florence, SAR+DEM (orange) varies less than SAR+AE (green) for the trainable backbones (CNN UNet, ImageNet-UNet), while the frozen-backbone SAR+AE configurations (TerraMind, DINOv3) show comparable variance. On Louisiana the spreads of the two signals are similar.}
\label{fig:seed_variance}
\end{figure*}

Each auxiliary-signal configuration was trained under three seeds (42, 7, 19). Two patterns emerge from the seed variance, shown in Figure~\ref{fig:seed_variance}.

First, DEM varies less across seeds than AE on Florence. The standard deviation of SAR+DEM Florence IoU ranges from $0.003$ to $0.009$ across the four backbones, while SAR+AE ranges from $0.003$ to $0.016$. The difference is driven by the two trainable backbones. CNN UNet SAR+AE and ImageNet-UNet SAR+AE have Florence standard deviations of $0.016$ and $0.015$, against $0.006$ and $0.003$ for their SAR+DEM counterparts. The two frozen-backbone SAR+AE configurations have lower variance, $0.005$ for TerraMind and $0.003$ for DINOv3, comparable to their DEM counterparts. The 64-dimensional AlphaEarth signal therefore introduces more seed sensitivity than the single-channel DEM, but mainly when paired with a fully trainable backbone.

Second, the wider AE spread coincides with higher AE means on Florence. The best individual Florence result for each backbone is a SAR+AE run, and the AlphaEarth mean exceeds the DEM mean on Florence for all four backbones. On Louisiana the standard deviations of the two signals are similar ($0.007$ to $0.016$ for both), and neither signal varies less than the other.

The epoch at which the best validation checkpoint was selected also varied with seed. Under seed 42, both trainable-backbone SAR+AE configurations selected epoch 1, before substantial training had occurred; under seeds 7 and 19 the same configurations selected later epochs (9 and 25 for CNN UNet, 28 and 18 for ImageNet-UNet) with higher test IoU. Among the SAR+DEM configurations, one run (DINOv3 SAR+DEM, seed 19) also selected epoch 1, while the rest selected epochs between 2 and 16. The epoch-1 selections were thus a seed-specific effect rather than a property of either auxiliary signal, and averaging over seeds removes their influence on the reported results.

\section{Discussion}

The findings show multiple important contributions. First, they show that single-temporal SAR segmentation benefits from external land-context information across all four backbone families, reinforcing the idea that the limiting factor is not only backbone architecture but the absence of a stable prior for interpreting ambiguous radar backscatter. Second, the comparison between AlphaEarth and DEM reveals a meaningful trade-off rather than a universal winner: AlphaEarth provides the strongest results on the harder Florence event and tends toward higher recall, while DEM is more stable across seeds and produces the strongest Louisiana result. Third, the event-level analysis demonstrates why aggregate validation metrics are insufficient for flood segmentation. Differences in flood prevalence, training-event similarity, and seed stability materially change the apparent ranking of configurations. The innovation of the paper is therefore the characterization of when auxiliary priors help, not merely the observation that adding AlphaEarth can improve IoU.

\subsection{Effect of AlphaEarth Embeddings Across Architectures}
\label{sec:ae-effect}

Averaged over three seeds, AlphaEarth embeddings improve flood segmentation over the SAR-only baseline for all four backbones on both test events. The improvement is larger on Florence, the harder cross-event test, than on Louisiana. On Florence, SAR+AE achieves the highest mean IoU of any configuration, DINOv3 at $0.078$, and the SAR+AE mean exceeds the SAR-only baseline for every backbone. This holds across backbones with very different pretraining, a from-scratch CNN UNet, an ImageNet-initialized UNet, a SAR-pretrained foundation model in TerraMind, and an optically-pretrained foundation model in DINOv3. The benefit of AlphaEarth is therefore not tied to any single pretraining regime.

The size of the AlphaEarth contribution interacts with the backbone. The two frozen-backbone foundation models show the largest and most seed-stable AlphaEarth gains on Florence, while the two fully-trainable backbones reach higher IoU under their best seeds but with greater seed-to-seed variance (Section~\ref{sec:seed-stability}). One interpretation is that a frozen backbone cannot adapt its own features to the test domain, so the externally supplied land-cover context contributes more consistently; a trainable backbone can in principle learn comparable context on its own, but does so with greater dependence on initialization. This is offered as an interpretation rather than a tested claim, since isolating the mechanism would require controlled experiments that hold capacity and trainability fixed.

\subsection{AlphaEarth compared to DEM}
\label{sec:ae-vs-dem}

\begin{figure*}[!t]
\centering
\includegraphics[width=\textwidth]{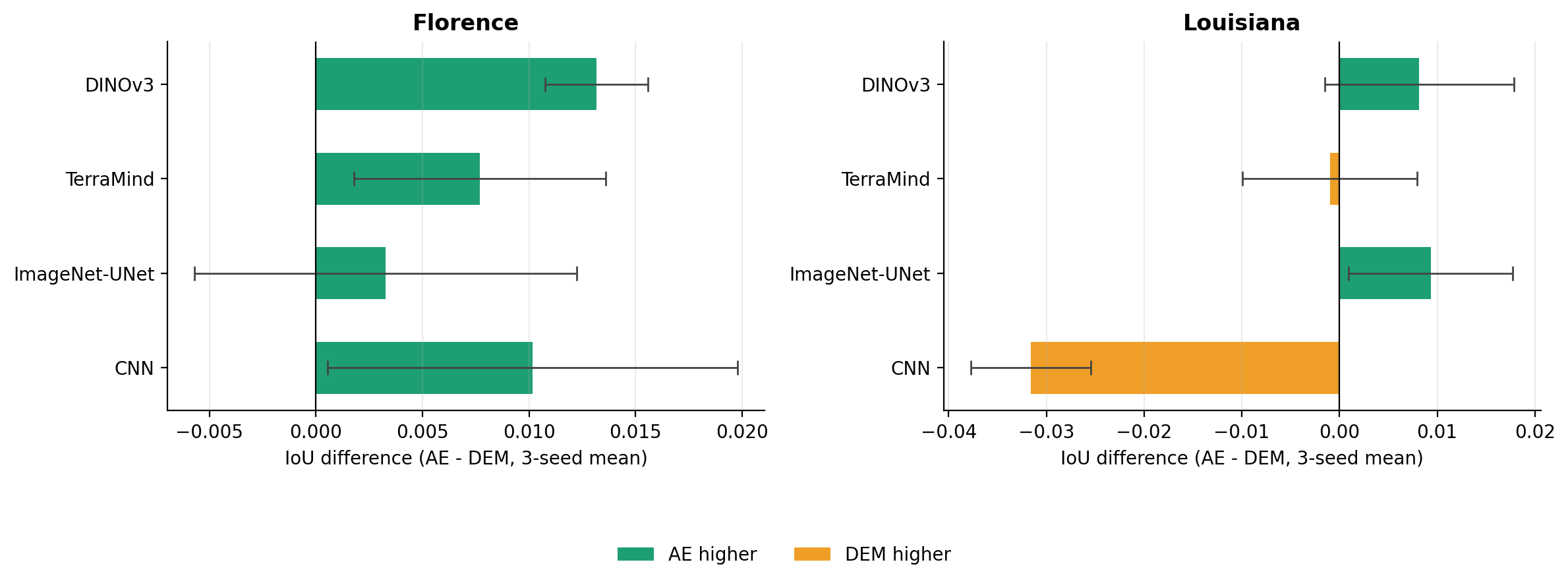}
\caption{Mean difference in test IoU between SAR+AE and SAR+DEM for each backbone and event, averaged over three seeds. Positive values indicate AlphaEarth exceeds DEM; whiskers show the standard error of the difference (over three seeds). AlphaEarth exceeds DEM for all four backbones on Florence, while results are mixed on Louisiana. Several per-backbone differences fall within one standard deviation of zero, indicating the direction of the effect is more consistent than its per-backbone magnitude.}
\label{fig:ae_minus_dem}
\end{figure*}

The two auxiliary signals differ in what they encode. DEM provides a single channel of absolute elevation, a static physical prior on where water collects. AlphaEarth provides a 64-dimensional learned embedding of land cover and surface context. Figure~\ref{fig:ae_minus_dem} shows the mean difference in test IoU between the two signals for each backbone and event.

On Florence, AlphaEarth exceeds DEM for all four backbones. The mean difference ranges from $0.003$ for ImageNet-UNet to $0.013$ for DINOv3. The direction of the effect is consistent across backbones, though the per-backbone margin is within one standard deviation of zero for ImageNet-UNet and TerraMind, so the size of the advantage is less certain than its direction. On Louisiana the comparison is mixed. AlphaEarth leads on ImageNet-UNet and DINOv3, DEM leads on CNN UNet, and the two are within one standard deviation on TerraMind. The single strongest Louisiana result is CNN UNet SAR+DEM at $0.198$, which pairs the simplest auxiliary signal with the only from-scratch backbone.

The two signals also occupy different operating points. AlphaEarth configurations tend toward higher recall and lower precision, while DEM configurations tend toward higher precision and lower recall. On Louisiana, ImageNet-UNet SAR+AE reaches recall $0.824$ at precision $0.164$, while CNN UNet SAR+DEM reaches precision $0.221$ at recall $0.658$ (seed 42). A learned land-cover embedding flags more candidate flood pixels, including correct detections that elevation alone misses, at the cost of more false positives. The preferable operating point depends on whether the downstream application is more sensitive to missed detections or to false alarms.

Across both events, AlphaEarth is the stronger signal on Florence, where it exceeds DEM for every backbone and produces the best single result, while on Louisiana the two signals are competitive and DEM produces the best single result. DEM is also a single elevation raster against a 64-dimensional embedding, a difference in input size that is relevant where the simpler signal is sufficient. The takeaway is therefore not that AlphaEarth is simply better than DEM: AlphaEarth is more useful under harder cross-event generalization and when a higher-recall operating point is required, whereas DEM is more stable across initializations, cheaper to store, and preferable when the simpler topographic prior is sufficient.

\subsection{Cross-Event Magnitude Differences}
\label{sec:cross-event}

Every configuration scores higher on Louisiana than on Florence. Two measured factors likely contribute substantially to this difference and are consistent with the observed IoU gap.

The first is flood-class prevalence. Flood pixels constitute $1.37\%$ of the Florence tiles and $5.12\%$ of the Louisiana tiles, a $3.7\times$ difference. IoU declines as the positive class becomes rarer, since a fixed number of false positives weighs more heavily against a smaller true-flood area. The per-tile distribution is more pronounced. The median Florence tile contains $0.17\%$ flood pixels against $0.44\%$ for Louisiana, so most Florence tiles are predominantly background. These tiles contribute false positives without true positives, reducing aggregate IoU. A substantial portion of the gap therefore reflects the lower flood prevalence of Florence rather than reduced model quality.

The second factor is each test event's similarity to the training distribution. The training set is dominated by Hurricane Harvey, which constitutes $73.6\%$ of training tiles and has $6.53\%$ prevalence. Louisiana ($5.12\%$) is closer to this value than Florence ($1.37\%$). The models were trained predominantly on higher-prevalence scenes and consequently transfer more effectively to the denser Louisiana event.

These prevalence values are consistent with the recorded characteristics of the two events. Florence flooded an extensive area of the Carolinas through combined coastal surge and river flooding, distributing inundation across many tiles. The August 2016 Louisiana event produced concentrated riverine flooding near Baton Rouge and Lafayette, yielding denser coverage per tile. We report these as contributing factors rather than a complete account, as the experiments do not separate prevalence, flood type, and training similarity.

\begin{figure}[!t]
\centering
\includegraphics[width=\columnwidth]{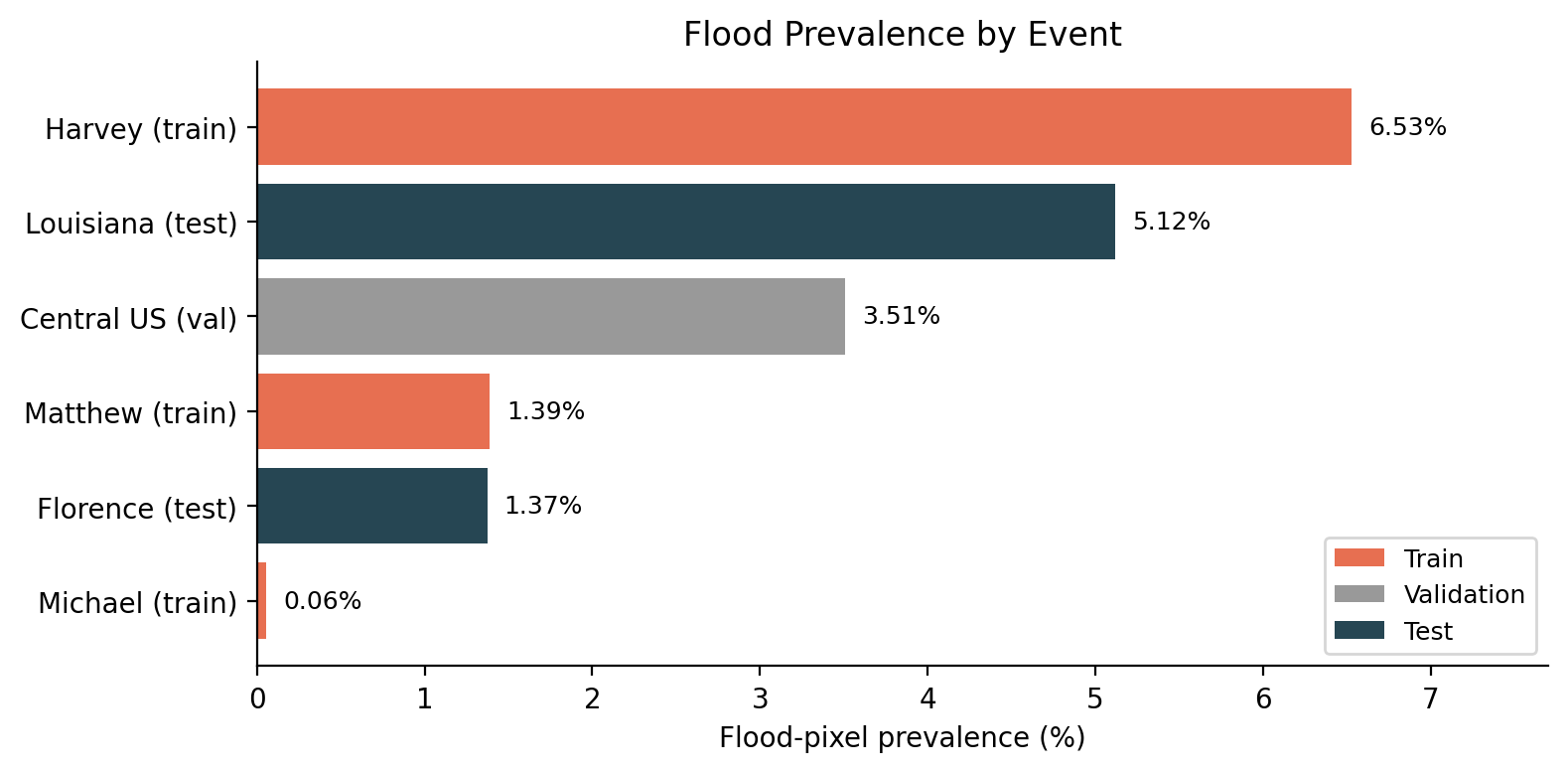}
\caption{Flood-pixel prevalence by event. The two test events differ significantly: Florence is $1.37\%$ flood, Louisiana is $5.12\%$. The training set is mostly Hurricane Harvey, which is $73.6\%$ of the training tiles and has $6.53\%$ flood.}
\label{fig:prevalence}
\end{figure}

\subsection{Training Behavior and the Generalization Gap}
\label{sec:convergence}

\begin{figure*}[!t]
\centering
\begin{subfigure}{0.48\textwidth}
    \includegraphics[width=\textwidth]{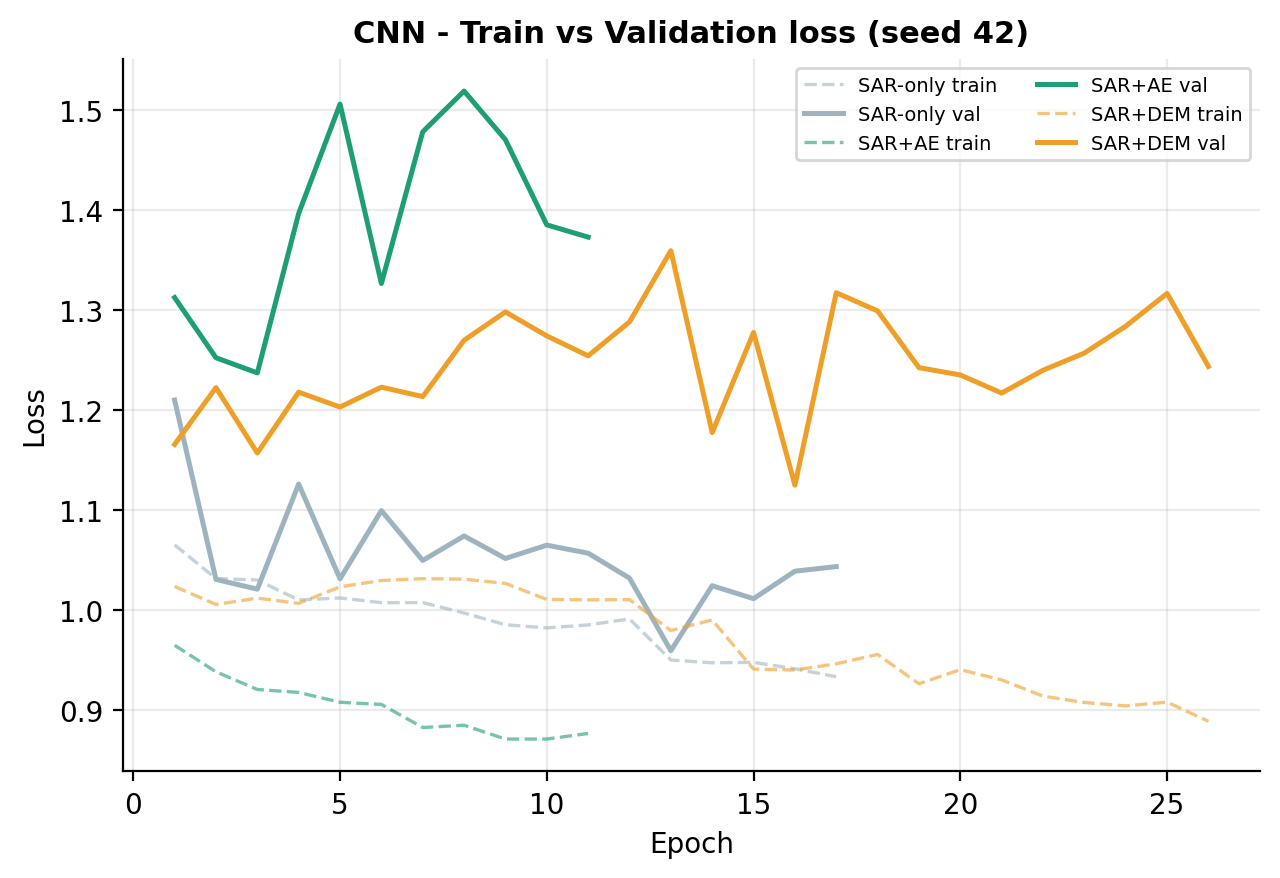}
    \caption{CNN UNet}
\end{subfigure}\hfill
\begin{subfigure}{0.48\textwidth}
    \includegraphics[width=\textwidth]{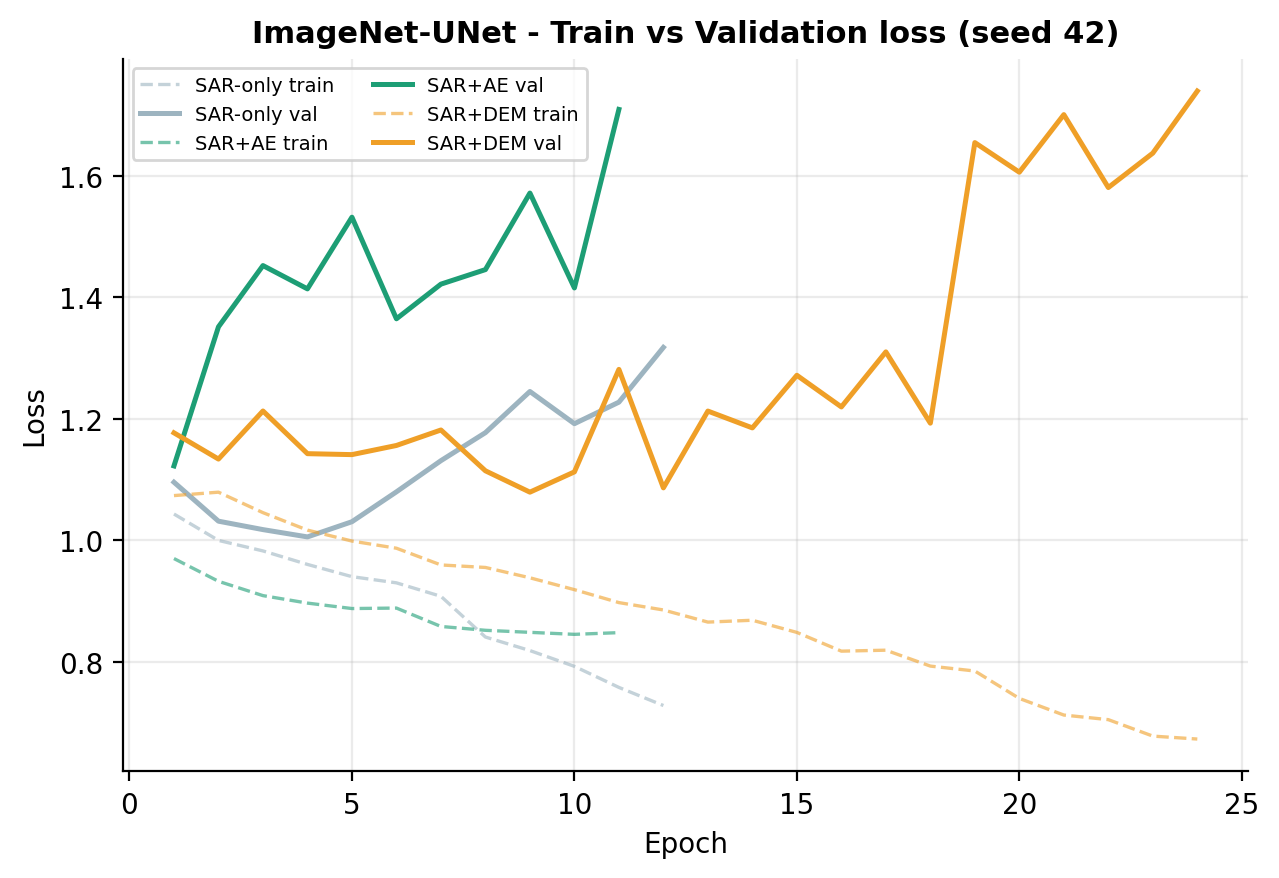}
    \caption{ImageNet-UNet}
\end{subfigure}

\vspace{0.6em}

\begin{subfigure}{0.48\textwidth}
    \includegraphics[width=\textwidth]{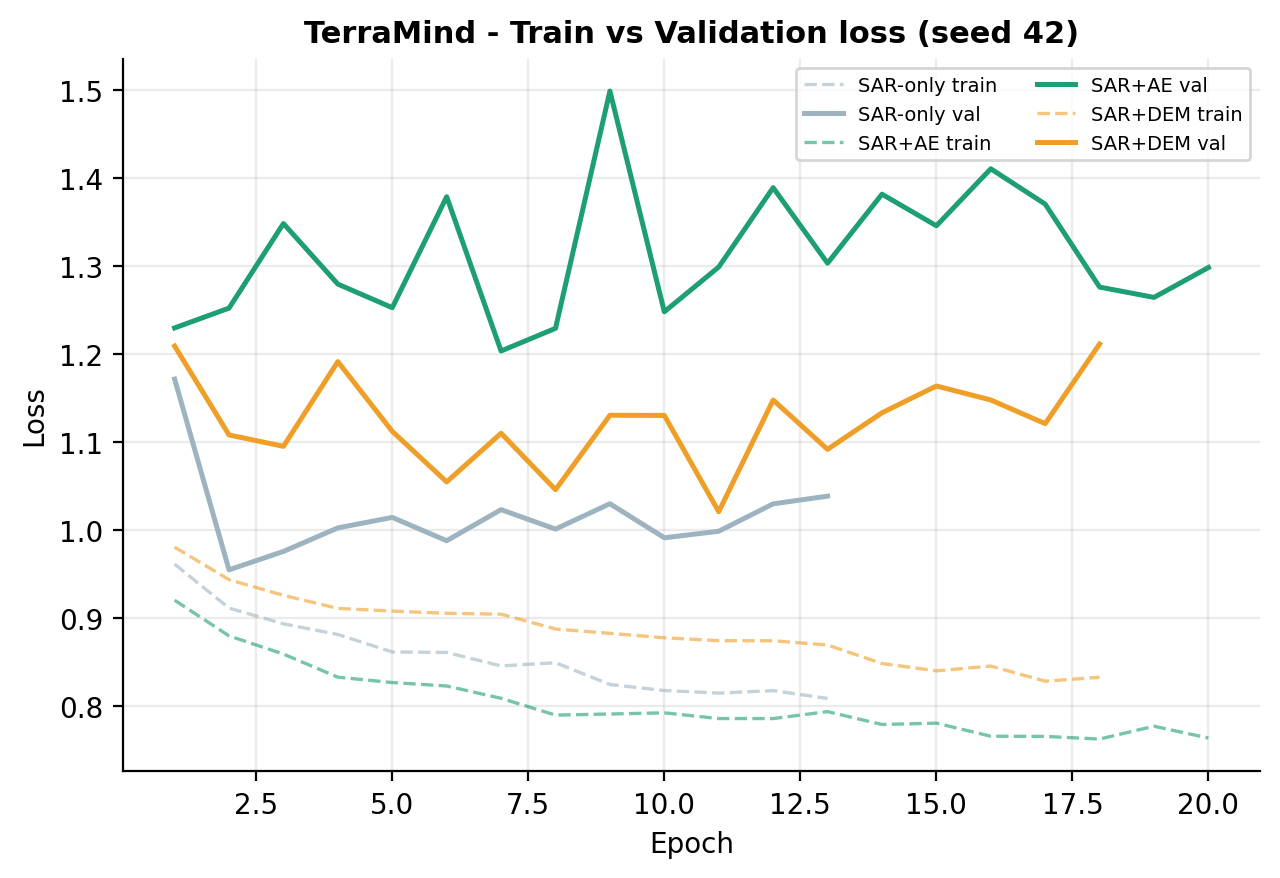}
    \caption{TerraMind}
\end{subfigure}\hfill
\begin{subfigure}{0.48\textwidth}
    \includegraphics[width=\textwidth]{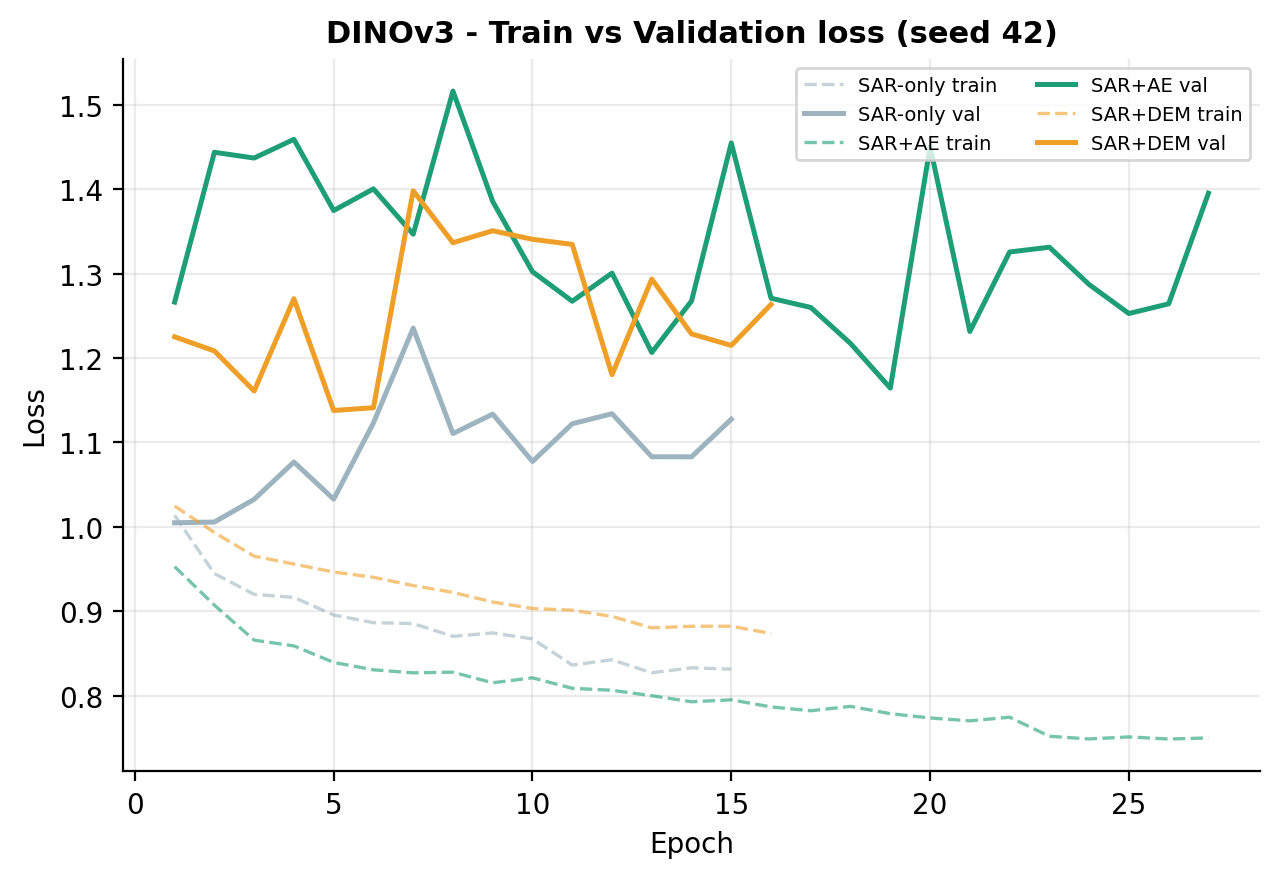}
    \caption{DINOv3}
\end{subfigure}

\caption{Training (dashed) and validation (solid) loss per backbone under seed 42. Across all four backbones, training loss decreases while validation loss rises or stays flat. The SAR+AE configurations show the widest train-validation gap, SAR+DEM a smaller gap, and SAR-only the smallest.}
\label{fig:loss_curves}
\end{figure*}

Figure~\ref{fig:loss_curves} shows train and validation loss per backbone under seed 42. A consistent pattern appears across all four backbones. Training loss (dashed) decreases steadily, while validation loss (solid) rises or remains flat. The models fit the training events but do not transfer that fit to the held-out validation event, which is the expected signature of an event-stratified split where training and validation come from different floods.

The size of this gap scales with the auxiliary signal. The SAR+AE configurations show the lowest training loss and the highest validation loss, producing the widest train-validation separation. SAR+DEM shows a smaller gap, and SAR-only the smallest. A richer auxiliary signal gives the model more capacity to fit patterns specific to the training events, and these patterns do not transfer to a different event. The $64$-dimensional AlphaEarth embedding provides more such capacity than the single-channel DEM, which is consistent with its wider gap and its higher seed variance (Section~\ref{sec:seed-stability}).

This pattern holds for the frozen foundation models as well as the fully-trainable backbones. TerraMind and DINOv3 train only a head and auxiliary branch, yet their validation loss rises across epochs in the same way. The generalization gap is therefore driven by the cross-event split and the auxiliary signal rather than by backbone trainability alone.

The curves explain two earlier observations. First, they show why model selection on validation IoU is necessary, since validation loss does not track test performance and continued training increases the gap. Second, they account for the low absolute IoU values, as the divergence between training and validation behavior reflects the difficulty of transferring a single post-event SAR fit across distinct flood events.

\subsection{Limitations}
\label{sec:limitations}

Several limitations bound the conclusions of this study. First, the ground-truth masks are Copernicus EMS rapid-mapping products rather than pixel-precise annotations. Their delineation error places a ceiling on achievable IoU that is independent of model quality, and we did not construct a manually verified subset to separate model error from label error.

Second, permanent water bodies were not masked. SAR returns low backscatter for both standing water and new inundation, so a portion of the detected flood area may correspond to rivers and lakes that are permanently wet rather than to flooding.

Third, the cross-event evaluation rests on two held-out events, Florence ($756$ tiles) and Louisiana ($336$ tiles). Two events are a narrow basis for general claims about cross-event transfer, and the conclusions may not extend to other flood types or regions. Additionally, all events are within the continental United States, so performance under different SAR acquisition conditions and flood regimes elsewhere is untested.

Fourth, seed coverage is uneven. The eight auxiliary-signal configurations were each trained in three seeds, but the four SAR-only baselines were trained in the primary seed only. Variance estimates also rest on three seeds, and we report means and standard deviations without formal significance testing, so the smaller per-backbone differences between AlphaEarth and DEM should be read as indicative rather than confirmed. Because the SAR-only baselines are single-seed, the improvement of each auxiliary prior over SAR-only is established only relative to the primary-seed baseline and its variance is not yet characterized.

Fifth, the AlphaEarth prior is a single 2024 annual layer applied uniformly across all events. As the 2024 reference year postdates every flood event from 2016 to 2018, the embedding cannot encode the target inundation, and the configuration is therefore free of event leakage. The trade-off is that the prior reflects 2024 land-surface conditions rather than those contemporaneous with each event, and sensitivity to alternative reference years was not evaluated.

Sixth, the study does not compare against an operational baseline such as simple backscatter thresholding, a permanent-water mask (for example JRC Global Surface Water), or a bi-temporal change-detection pipeline. The auxiliary priors are therefore shown to improve a single-temporal segmentation model, but not to match or exceed established operational alternatives, and permanent-water disambiguation is motivated more strongly than it is directly demonstrated.

These boundaries suggest a clear future-research agenda. The findings are limited to CONUS events and only two held-out test floods, so the observed AE-versus-DEM trade-off may not generalize to other hydrologic regimes, SAR acquisition geometries, land-cover distributions, or international mapping contexts. Future work should prioritize full multi-seed coverage of all configurations including the SAR-only baselines, leave-one-event-out or event-balanced cross-validation, a permanent-water-specific error analysis, an analysis of alternative AE reference years, and evaluation against simple operational baselines such as thresholding, permanent-water masks, or bi-temporal change detection where available. These additions would transform the current evidence from a strong controlled study into a more definitive operational benchmark.

\section{Concluding Remarks}
\label{sec:conclusion}

This study presents a controlled evaluation of auxiliary land-context signals for single-temporal post-event SAR flood segmentation, comparing learned land-cover embeddings (AlphaEarth) against a simple topographic prior (DEM) across four SAR backbones. The contributions are threefold. First, a controlled comparison of four backbones spanning four pretraining regimes, from-scratch, ImageNet-pretrained, SAR-pretrained, and optical-satellite-pretrained, each evaluated in SAR-only, SAR+DEM, and SAR+AE configurations under an identical fusion design, training protocol, and event-stratified split with per-event test reporting. Second, the finding that both auxiliary signals improve on the SAR-only baseline across all backbones, with AlphaEarth exceeding DEM on the harder cross-event test (Florence) for every backbone while the two are competitive on the easier event (Louisiana), where DEM yields the single best result. Third, a seed-stability analysis showing that DEM is the more stable signal across random seeds while AlphaEarth reaches higher peak performance on the harder event, and that a clean architecture-dependent pattern visible under a single seed does not survive averaging over three seeds.

Scientifically, the novelty of this work is not a new segmentation architecture but the systematic isolation of how prior information changes model behavior across four backbone families: a from-scratch CNN, an ImageNet-pretrained CNN, a SAR-pretrained geospatial foundation model, and an optical-satellite-pretrained foundation model. By adding DEM as a minimal physical baseline and evaluating AlphaEarth as a richer learned land-surface prior under the same fusion design, optimizer, loss, split, and evaluation protocol, the study provides evidence about the relative value, stability, and precision-recall trade-offs of these two forms of auxiliary context. The per-event and multi-seed analyses are central because they show that a single headline IoU would obscure the main insight: AlphaEarth is most advantageous on the harder held-out event, whereas DEM is more stable and can be stronger when topography is sufficient.

From a practical perspective, the results point toward a deployable strategy under the constraints emergency-response agencies actually face. A clean pre-event SAR acquisition may be unavailable, poorly registered, or environmentally mismatched, while waiting for additional imagery reduces operational value. A lightweight auxiliary branch that accepts a readily available prior can be attached to multiple SAR backbones without redesigning the full model. Practitioners should not treat auxiliary context as a single choice: DEM may be preferred when stability, storage, cost, and conservative precision are priorities, while AlphaEarth may be more useful when the objective is to increase recall or improve transfer to difficult, low-prevalence events. This turns the method from a purely academic comparison into a decision framework for selecting auxiliary priors according to event type, risk tolerance, and computational constraints.

A lightweight auxiliary branch fused into an existing SAR backbone is a low-cost, architecture-agnostic way to improve single-temporal post-event flood segmentation. The choice between signals reflects a trade-off rather than a clear winner. AlphaEarth provides the larger benefit where cross-event generalization is hardest, at the cost of greater seed sensitivity and a $64$-dimensional input, while DEM is more stable, cheaper to store, and stronger on the easier event. Future work could address the main limitations of this study: pixel-precise labels to separate model error from label error, a permanent-water mask to isolate new inundation, a larger and more geographically diverse set of test events, and full multi-seed coverage of all configurations. Evaluating the two signals in combination, and across regions beyond the continental United States, would further clarify when a learned land-cover prior is preferable to a simple topographic one. We frame AlphaEarth/DEM fusion not as a demonstrated replacement for bi-temporal change detection, but as a practical single-temporal fallback or complement for when a registered, seasonally comparable pre-event SAR acquisition is unavailable or unreliable.

A practical implementation pathway would begin with a modular flood-mapping pipeline that ingests post-event Sentinel-1 RTC imagery, retrieves a co-registered DEM and AlphaEarth tile for the affected area, applies the normalization steps defined in this study, and runs a calibrated SAR+prior segmentation model. For deployment, the pipeline should expose both a conservative DEM-informed mode and a higher-recall AlphaEarth-informed mode, allowing analysts to select the operating point according to the cost of missed inundation versus false alarms.

\section*{Data Availability}
The data supporting the findings of this study are derived from two publicly available sources. The ImpactMesh-Flood dataset, which includes the post-event Sentinel-1 RTC imagery and the corresponding flood extent masks, is accessible through \href{https://huggingface.co/datasets/ibm-esa-geospatial/ImpactMesh-Flood}{Hugging Face}. Additionally, the AlphaEarth embeddings are available via Google Earth Engine.

\section*{Code Availability}
The code that supports the findings of this study is available from the corresponding author upon request.

\section*{Acknowledgment}
During the preparation of this work, the authors used Claude (Anthropic) to assist with
code generation, language editing, and LaTeX formatting. The authors reviewed and edited
all such content and take full responsibility for the content of the article.

 
\appendices
\section{Per-Seed Results}
\label{sec:appendix}

\begin{table*}[!t]
\centering
\small
\setlength{\tabcolsep}{4pt}
\caption{Per-seed test results on the Florence event (EMSR311, $n=756$). Flood-class IoU, F1, precision (P), and recall (R) for every configuration and seed. Ep is the epoch of the selected best-validation checkpoint. SAR-only baselines are single-seed.}
\label{tab:appendix_florence}
\begin{tabular}{llcccccc}
\hline
\textbf{Backbone} & \textbf{Config} & \textbf{Seed} & \textbf{Ep} & \textbf{IoU} & \textbf{F1} & \textbf{P} & \textbf{R} \\
\hline
CNN & SAR-only & 42 & 7 & 0.0406 & 0.0781 & 0.0441 & 0.3399 \\
 & SAR+DEM & 7 & 2 & 0.0426 & 0.0818 & 0.0503 & 0.2185 \\
 & SAR+DEM & 19 & 2 & 0.0418 & 0.0803 & 0.0446 & 0.3990 \\
 & SAR+DEM & 42 & 16 & 0.0522 & 0.0992 & 0.0572 & 0.3755 \\
 & SAR+AE & 7 & 9 & 0.0455 & 0.0870 & 0.1649 & 0.0591 \\
 & SAR+AE & 19 & 25 & 0.0737 & 0.1372 & 0.1418 & 0.1329 \\
 & SAR+AE & 42 & 1 & 0.0479 & 0.0913 & 0.0699 & 0.1319 \\
\hline
ImageNet-UNet & SAR-only & 42 & 2 & 0.0567 & 0.1074 & 0.0667 & 0.2754 \\
 & SAR+DEM & 7 & 12 & 0.0603 & 0.1138 & 0.0715 & 0.2788 \\
 & SAR+DEM & 19 & 8 & 0.0544 & 0.1032 & 0.0600 & 0.3668 \\
 & SAR+DEM & 42 & 14 & 0.0580 & 0.1097 & 0.0715 & 0.2356 \\
 & SAR+AE & 7 & 28 & 0.0645 & 0.1212 & 0.1360 & 0.1093 \\
 & SAR+AE & 19 & 18 & 0.0739 & 0.1377 & 0.1385 & 0.1369 \\
 & SAR+AE & 42 & 1 & 0.0441 & 0.0844 & 0.0578 & 0.1561 \\
\hline
TerraMind & SAR-only & 42 & 3 & 0.0463 & 0.0885 & 0.0499 & 0.3904 \\
 & SAR+DEM & 7 & 6 & 0.0721 & 0.1346 & 0.0822 & 0.3714 \\
 & SAR+DEM & 19 & 4 & 0.0658 & 0.1234 & 0.0746 & 0.3566 \\
 & SAR+DEM & 42 & 8 & 0.0549 & 0.1041 & 0.0597 & 0.4065 \\
 & SAR+AE & 7 & 3 & 0.0714 & 0.1332 & 0.1045 & 0.1838 \\
 & SAR+AE & 19 & 2 & 0.0776 & 0.1440 & 0.1243 & 0.1709 \\
 & SAR+AE & 42 & 10 & 0.0669 & 0.1254 & 0.1126 & 0.1414 \\
\hline
DINOv3 & SAR-only & 42 & 5 & 0.0446 & 0.0854 & 0.0486 & 0.3533 \\
 & SAR+DEM & 7 & 9 & 0.0658 & 0.1235 & 0.0776 & 0.3025 \\
 & SAR+DEM & 19 & 1 & 0.0673 & 0.1261 & 0.0827 & 0.2652 \\
 & SAR+DEM & 42 & 6 & 0.0623 & 0.1173 & 0.0682 & 0.4175 \\
 & SAR+AE & 7 & 10 & 0.0821 & 0.1517 & 0.1478 & 0.1558 \\
 & SAR+AE & 19 & 5 & 0.0765 & 0.1420 & 0.1551 & 0.1310 \\
 & SAR+AE & 42 & 17 & 0.0763 & 0.1418 & 0.1236 & 0.1663 \\
\hline
\end{tabular}
\end{table*}

\begin{table*}[!t]
\centering
\small
\setlength{\tabcolsep}{4pt}
\caption{Per-seed test results on the Louisiana event (EMSR176, $n=336$). Flood-class IoU, F1, precision (P), and recall (R) for every configuration and seed. Ep is the epoch of the selected best-validation checkpoint. SAR-only baselines are single-seed.}
\label{tab:appendix_louisiana}
\begin{tabular}{llcccccc}
\hline
\textbf{Backbone} & \textbf{Config} & \textbf{Seed} & \textbf{Ep} & \textbf{IoU} & \textbf{F1} & \textbf{P} & \textbf{R} \\
\hline
CNN & SAR-only & 42 & 7 & 0.1033 & 0.1873 & 0.1241 & 0.3821 \\
 & SAR+DEM & 7 & 2 & 0.2062 & 0.3419 & 0.2979 & 0.4012 \\
 & SAR+DEM & 19 & 2 & 0.1908 & 0.3204 & 0.2058 & 0.7237 \\
 & SAR+DEM & 42 & 16 & 0.1980 & 0.3305 & 0.2206 & 0.6584 \\
 & SAR+AE & 7 & 9 & 0.1696 & 0.2899 & 0.2079 & 0.4789 \\
 & SAR+AE & 19 & 25 & 0.1584 & 0.2734 & 0.1878 & 0.5024 \\
 & SAR+AE & 42 & 1 & 0.1722 & 0.2938 & 0.1852 & 0.7109 \\
\hline
ImageNet-UNet & SAR-only & 42 & 2 & 0.1357 & 0.2390 & 0.1603 & 0.4697 \\
 & SAR+DEM & 7 & 12 & 0.1636 & 0.2812 & 0.1742 & 0.7292 \\
 & SAR+DEM & 19 & 8 & 0.1410 & 0.2471 & 0.1571 & 0.5778 \\
 & SAR+DEM & 42 & 14 & 0.1624 & 0.2794 & 0.1722 & 0.7397 \\
 & SAR+AE & 7 & 28 & 0.1720 & 0.2935 & 0.1868 & 0.6851 \\
 & SAR+AE & 19 & 18 & 0.1650 & 0.2832 & 0.1784 & 0.6869 \\
 & SAR+AE & 42 & 1 & 0.1580 & 0.2729 & 0.1635 & 0.8242 \\
\hline
TerraMind & SAR-only & 42 & 3 & 0.1437 & 0.2514 & 0.1548 & 0.6681 \\
 & SAR+DEM & 7 & 6 & 0.1755 & 0.2986 & 0.1970 & 0.6165 \\
 & SAR+DEM & 19 & 4 & 0.1531 & 0.2656 & 0.1713 & 0.5904 \\
 & SAR+DEM & 42 & 8 & 0.1585 & 0.2736 & 0.1699 & 0.7014 \\
 & SAR+AE & 7 & 3 & 0.1497 & 0.2604 & 0.1565 & 0.7756 \\
 & SAR+AE & 19 & 2 & 0.1666 & 0.2856 & 0.1770 & 0.7388 \\
 & SAR+AE & 42 & 10 & 0.1679 & 0.2875 & 0.2018 & 0.4994 \\
\hline
DINOv3 & SAR-only & 42 & 5 & 0.1326 & 0.2341 & 0.1511 & 0.5195 \\
 & SAR+DEM & 7 & 9 & 0.1803 & 0.3055 & 0.2031 & 0.6160 \\
 & SAR+DEM & 19 & 1 & 0.1635 & 0.2811 & 0.2032 & 0.4560 \\
 & SAR+DEM & 42 & 6 & 0.1488 & 0.2590 & 0.1642 & 0.6132 \\
 & SAR+AE & 7 & 10 & 0.1720 & 0.2936 & 0.1787 & 0.8212 \\
 & SAR+AE & 19 & 5 & 0.1668 & 0.2859 & 0.1755 & 0.7716 \\
 & SAR+AE & 42 & 17 & 0.1783 & 0.3027 & 0.1920 & 0.7139 \\
\hline
\end{tabular}
\end{table*}

\clearpage  
\bibliographystyle{IEEEtran}
\bibliography{ref}

\begin{IEEEbiographynophoto}{Sanjay Thasma}
received the B.S. degree in Computer Science from the University of
Wisconsin--Madison, Madison, WI, USA. He is currently pursuing the M.S. degree
in Computer Science at Texas A\&M University, College Station, TX, USA.
\end{IEEEbiographynophoto}

\begin{IEEEbiographynophoto}{Yu-Hsuan Ho}
received the B.S. degree in Civil Engineering and the M.S. degree in
Transportation Engineering from National Taiwan University, Taipei, Taiwan. She
is currently pursuing the Ph.D. degree in the Zachry Department of Civil and
Environmental Engineering at Texas A\&M University, College Station, TX, USA. Her
research interests include computer vision and urban data science. In her free
time, she enjoys watching movies and visiting film festivals.
\end{IEEEbiographynophoto}

\begin{IEEEbiographynophoto}{Ali Mostafavi}
received the Ph.D. degree in Civil Engineering from Purdue University, West
Lafayette, IN, USA, in 2013. He is currently a Professor in the Zachry
Department of Civil and Environmental Engineering, Texas A\&M University, College
Station, TX, USA, where he holds the Zachry Professorship in Design and
Construction Integration II and a joint appointment in the Department of Computer
Science and Engineering. He directs the UrbanResilience.AI Lab.
\end{IEEEbiographynophoto}
\end{document}